\documentclass{article}

\usepackage[preprint]{neurips_2025}

\usepackage[utf8]{inputenc} 
\usepackage[T1]{fontenc}    
\usepackage{hyperref}       
\usepackage{url}            
\usepackage{booktabs}       
\usepackage{amsfonts}       
\usepackage{nicefrac}       
\usepackage{microtype}      
\usepackage{xcolor}         
\usepackage{graphicx}
\usepackage{cleveref} 

\title{Rubric-Conditioned LLM Grading: Alignment, Uncertainty, and Robustness}

\author{%
  Haotian Deng \\
  Purdue University\\
  West Lafayette, IN 47906 \\
  \texttt{deng254@purdue.edu} \\
  \And
  Chris Farber \\
  Purdue University\\
  West Lafayette, IN 47906 \\
  \texttt{cfarber@purdue.edu} \\
  \And
  Jiyoon Lee \\
  Purdue University\\
  West Lafayette, IN 47906 \\
  \texttt{lee4862@purdue.edu} \\
  \And
  David Tang \\
  Purdue University\\
  West Lafayette, IN 47906 \\
  \texttt{tang627@purdue.edu} \\
}

\begin{document}

\maketitle

\begin{abstract}
Automated short-answer grading (ASAG) remains a challenging task due to the
linguistic variability of student responses and the need for nuanced,
rubric-aligned partial credit. While Large Language Models (LLMs) offer a
promising solution, their reliability as automated judges in rubric-based
settings requires rigorous assessment. In this paper, we systematically
evaluate the performance of LLM-judges for rubric-based short-answer grading.
We investigate three key aspects: the alignment of LLM grading with expert
judgment across varying rubric complexities, the trade-off between uncertainty
and accuracy facilitated by a consensus-based deferral mechanism, and the
model's robustness under random input perturbations and adversarial attacks.
Using the SciEntsBank benchmark and Qwen 2.5-72B, we find that alignment is
strong for binary tasks but degrades with increased rubric granularity. Our
``Trust Curve'' analysis demonstrates a clear trade-off where filtering
low-confidence predictions improves accuracy on the remaining subset.
Additionally, robustness experiments reveal that while the model is resilient
to prompt injection, it is sensitive to synonym substitutions. Our work
provides critical insights into the capabilities and limitations of
rubric-conditioned LLM judges, highlighting the importance of uncertainty
estimation and robustness testing for reliable deployment.
\end{abstract}

\section{Introduction}

Short-answer grading shows up everywhere –- university courses, online
education, and compliance or safety checks –- yet it remains hard to automate.
Free-form responses vary in phrasing and partial correctness; graders care about
specific rubric criteria (key ideas, justification, units, reasoning steps), not
just surface similarity. Hand-crafted rules rarely cover the long tail of valid
answers, and fully manual grading does not scale. At the same time, collecting
large, task-specific labeled datasets is expensive and often impossible. Recent
“LLM-as-a-judge” approaches are promising, but out of the box, they can be
unreliable, inconsistent, or brittle to harmless rewordings, especially when we
only have the rubric and a small number of graded examples.

\paragraph{Research Question:} We study the rubric-only (or low-label) grading
setting: Given a textual rubric and many free-form responses, to what extent can
we automatically assign reliable scores with minimal human labeling?  We
decompose this into three specific research questions (RQs):

\begin{itemize}
    \item \textbf{RQ1 (Alignment):} To what extent does the model align with expert judgments on rubric criteria, and how does this alignment vary with rubric complexity?
    \item \textbf{RQ2 (Uncertainty):} Can we improve system reliability by identifying uncertain predictions and selectively withdrawing them?
    \item \textbf{RQ3 (Robustness):} How stable is the grader under paraphrases, minor perturbations, and adversarial attacks?
\end{itemize}

This setting is relatively underexplored in current evaluation work. We
investigate a systematic and pragmatic approach that treats the rubric as the
primary supervision signal, uses a small number of graded anchors for light
calibration, and includes an explicit option to defer to a human when confidence
is low. Our goal is not to replace human graders but to understand how far
rubric-conditioned judging can go under realistic data constraints.

\section{Related Work}

\paragraph{Automated Short Answer Grading (ASAG).} Traditional ASAG methods
relied on feature engineering or fine-tuning BERT-based models, which required
collecting hundreds of labeled examples for every new question to achieve high
accuracy \citep{IntJArtifIntellEduc/BurrowsI15, devlin_bert_2019}.

\paragraph{LLM-as-a-judge (LAJ)} recently emerged as a line of research in
natural language processing and machine learning \citep{guSurveyLLMasaJudge2025,
liLLMsasJudgesComprehensiveSurvey2024}. This paradigm uses LLMs to evaluate
outputs by following specific instructions or criteria, eliminating the need for
large labeled training datasets \citep{guSurveyLLMasaJudge2025}.
\citet{Proc.1stInt.WorkshopLargeLang.ModelsCode/GrandelD24} demonstrate the
utility of LLMs like GPT-4 for enhancing assessment in parallel functional
programming courses, focusing on improving grading efficiency and objectivity.
\citet{BMCMedEduc/Grevisse24} similarly explores the potential of LLMs for
grading in medical education. Others have explored integrating LLMs into
human-in-the-loop workflows to refine feedback \citep{chu_llm-based_2025,
yang_pensieve_2025} or utilizing Retrieval-Augmented Generation (RAG) to enhance
grading accuracy \citep{qiu_stella_2025}. While these works focus on
domain-specific applications and efficiency with human feedback, our work
addresses the safety and robustness required for autonomous deployment. We
systematically test adversarial resilience against prompt injections and
introduce a consensus-based deferral mechanism to actively mitigate
hallucination risks, pushing the boundary of reliable autograding without direct
human intervention.

\subsection{Improving LLM Judges} Several lines of work aim to increase
accuracy, robustness, and stability of LAJ by intervening at three layers – how
the model is prompted, how it is trained, and how outputs are aggregated.

\paragraph{Prompting.} Chain-of-Thought (CoT) prompting elicits explicit
intermediate reasoning, often improving robustness and transparency on
reasoning-heavy tasks such as mathematics relative to standard short-answer
prompting \citep{weiChainofThoughtPromptingElicits}. Notably, automatic
chain-of-thought achieves comparable performance to manual CoT.
\citep{zhangAutomaticChainThought2022} Pairwise evaluation is shown to have
significant improvement over direct scoring and is more in agreement with human
judgment. \citep{liuAligningHumanJudgement2025}

\paragraph{Training.} \citet{huangThinkJLearningThink2025} investigate
fine-tuning LAJ via reinforcement learning on reasoning traces -— using both
offline and online RL -– to strengthen consistency and deliberative quality in
judge outputs \citep{huangThinkJLearningThink2025}.

\paragraph{Aggregation.} Multi-run or multi-model aggregation can reduce
variance by averaging or reconciling diverse signals. While not a judge per se,
MAATS demonstrates the value of multi-LLM feedback loops in translation,
combining multiple agents and MQM-style evaluation to improve downstream quality
\citep{wangMAATSMultiAgentAutomated2025}. Relatedly, broader LLM evaluation work
has explored experimental designs that aggregate outputs across samples to
stabilize estimates, a pattern that can inform LAJ evaluation setup
\citep{huangHumanityConversationalAI2024,
mohammadiEvalMORAALInterpretableChainofThought2025}. Our consensus mechanism
draws inspiration from Self-Consistency \citep{wang_self-consistency_2023},
which demonstrates that aggregating multiple sampling paths improves reasoning
performance. We extend this by using agreement not just to improve the answer,
but as a proxy for confidence to trigger human deferral.

\subsection{Evaluating LLM Judges}

\paragraph{Correlation to Human Judgment.} A central question for LAJ is how
closely model judgments align with expert labels.
\citet{hanJudgesVerdictComprehensive2025} conduct a large-scale cross-evaluation
of 54 LLM judges on expert-labeled datasets and report both correlation-based
agreement and Cohen's $\kappa$ as primary criteria
\citep{hanJudgesVerdictComprehensive2025}. \citet{sajuFactsAreHarder2025} study
multilingual, multi-topic fact-checking and find that LLM judges show limited
accuracy and a high no-response rate, especially outside high-resource languages
and familiar topics \citep{sajuFactsAreHarder2025}. 

\paragraph{Adversarial Robustness.} Recent studies also probe the robustness of
LAJ under adversarial and safety-critical conditions.
\citet{zhengCheatingAutomaticLLM2025} show that “null models” producing constant
outputs can secure surprisingly high win rates on some automatic judge
benchmarks, revealing brittleness in pairwise-comparison-style evaluation
pipelines \citep{zhengCheatingAutomaticLLM2025}.
\citet{liuGoalOrientedPromptAttack2023} propose goal-oriented prompt attacks
that elicit harmful behaviors with high success rates, indicating that judge
prompts and scoring rules can be exploited
\citep{liuGoalOrientedPromptAttack2023}. Complementing these findings,
\citet{zhangJADELinguisticsbasedSafety2023} introduce JADE, a linguistics-based
safety evaluation platform, and demonstrate that simple linguistic perturbations
(including random mutations) can circumvent safety guardrails and induce
responses to dangerous prompts \citep{zhangJADELinguisticsbasedSafety2023}.
Collectively, these results argue for robustness-aware judge design and for
adversarially resilient evaluation protocols.

\section{Methods}

\subsection{Core Approach} We developed a rubric-conditioned grading pipeline
using Large Language Models to evaluate student short answers. Unlike
traditional supervised methods that require fine-tuning, our approach treats the
rubric as the primary supervision signal. To systematically evaluate
reliability, we employ a multi-faceted evaluation framework corresponding to our
three research questions. First, we assess \textbf{alignment} by comparing model
grades against expert labels across varying rubric complexities (2-way, 3-way,
and 5-way). Second, to manage \textbf{uncertainty}, we introduce a
consensus-based deferral mechanism that queries the model $N$ times, assigning a
grade only if a sufficient majority agrees; this allows us to optimize the
trade-off between automation coverage and grading reliability. Finally, we
evaluate \textbf{robustness} by subjecting the model to linguistic perturbations
and adversarial ``null model'' attacks, testing its resilience to surface-level
changes and malicious inputs.

\subsection{Technical Details}

\paragraph{Model.} We utilize Qwen 2.5 (72B Instruct), an open-weights model
served via the Purdue RCAC GenAI API. We selected this model for its strong
reasoning capabilities and accessibility, serving as a representative baseline
for state-of-the-art open models. No additional fine-tuning was performed.

\paragraph{Dataset.} We utilize the SciEntsBank dataset
\citep{NkaziSciEntsBankDatasets2025}, a benchmark for short-answer grading. It
contains scientific questions, reference answers, and student responses labeled
with 2-way (Correct/Incorrect), 3-way, and 5-way grading schemes.

\paragraph{Prompt Design.} We construct prompts that explicitly include the
question, reference answer, and the specific grading rubric guidelines provided
by the dataset. (See Appendix A for the full prompt template).

\paragraph{Implementation.} We built a custom evaluation harness that supports
batch processing, structured logging, and extensive data augmentation using the
\texttt{nlpaug} library for robustness testing.\footnote{Source code and all results presented in this paper can be found at \url{https://github.com/PROgram52bc/CS577_llm_judge}.}

\section{Experiments}

\subsection{Error Analysis (Confusion Matrix)} 
\label{sec:exp_confusion}

\begin{figure}[htbp]
  \centering
  \includegraphics[width=0.8\textwidth]{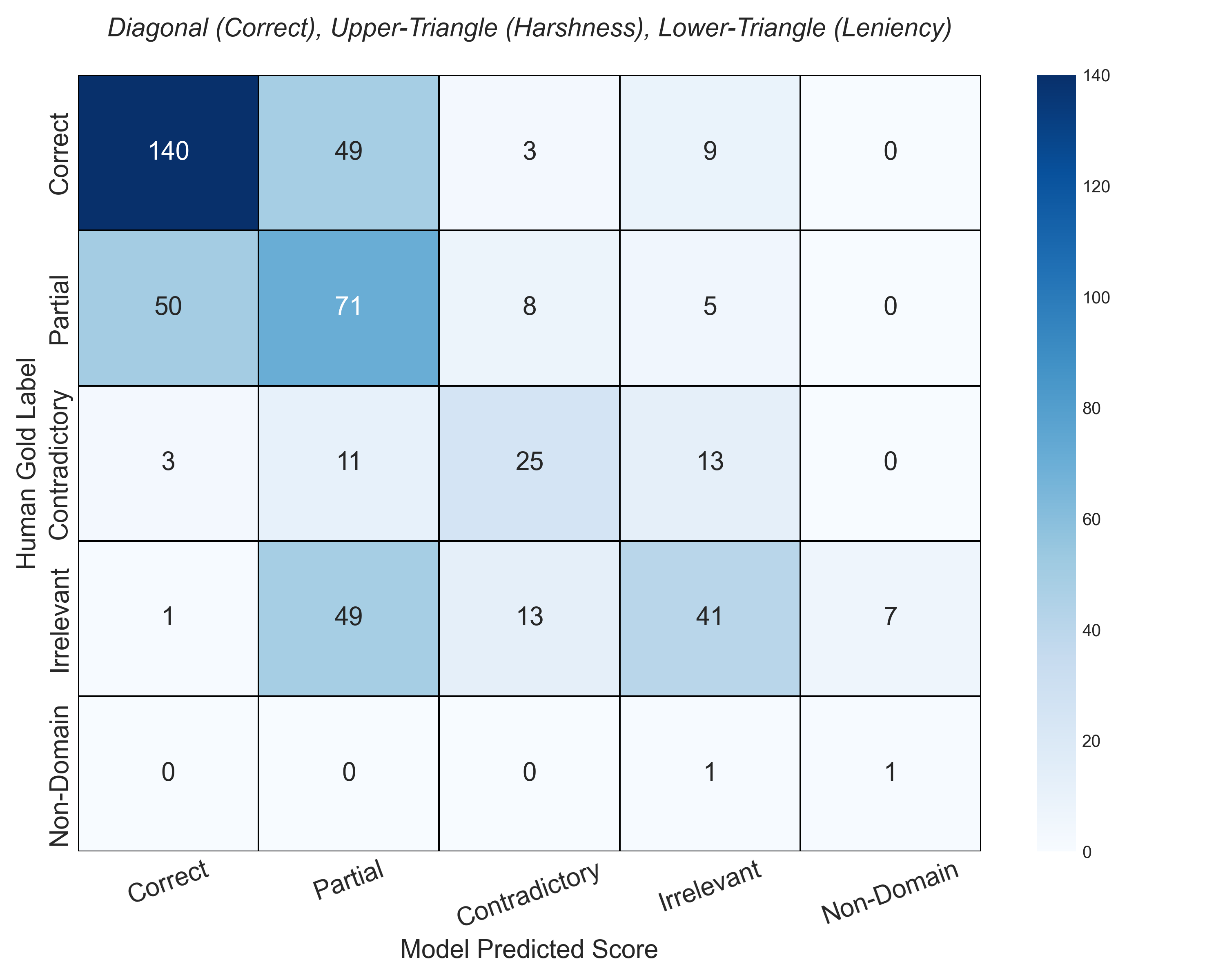}
  \caption{Confusion matrix comparing model predictions to human gold labels. The diagonal represents correct classifications, while the upper and lower triangles indicate where the model was harsher or more lenient than the human graders, respectively.}
  \label{fig:confusion}
\end{figure}

To investigate RQ1 (Alignment), we evaluated the 5-way grading scheme to
identify specific biases in model predictions compared to human labels.  The
confusion matrix in \Cref{fig:confusion} reveals that the model is generally
more lenient than human graders/reference labels. The most common error type was
classifying ``Partially Correct'' answers as ``Correct,'' and vice versa. The
model also frequently classifies ``Irrelevant'' answers as ``Partially
Correct,'' indicating a lack of nuance in awarding partial credit.

\subsection{Label Scheme Complexity} 
\label{sec:exp_complexity}
Also addressing RQ1, we
compared performance across 2-way (Correct/Incorrect), 3-way, and 5-way grading
schemes to measure how alignment degrades with rubric complexity.

\begin{figure}[htbp]
  \centering
  \includegraphics[width=0.9\textwidth]{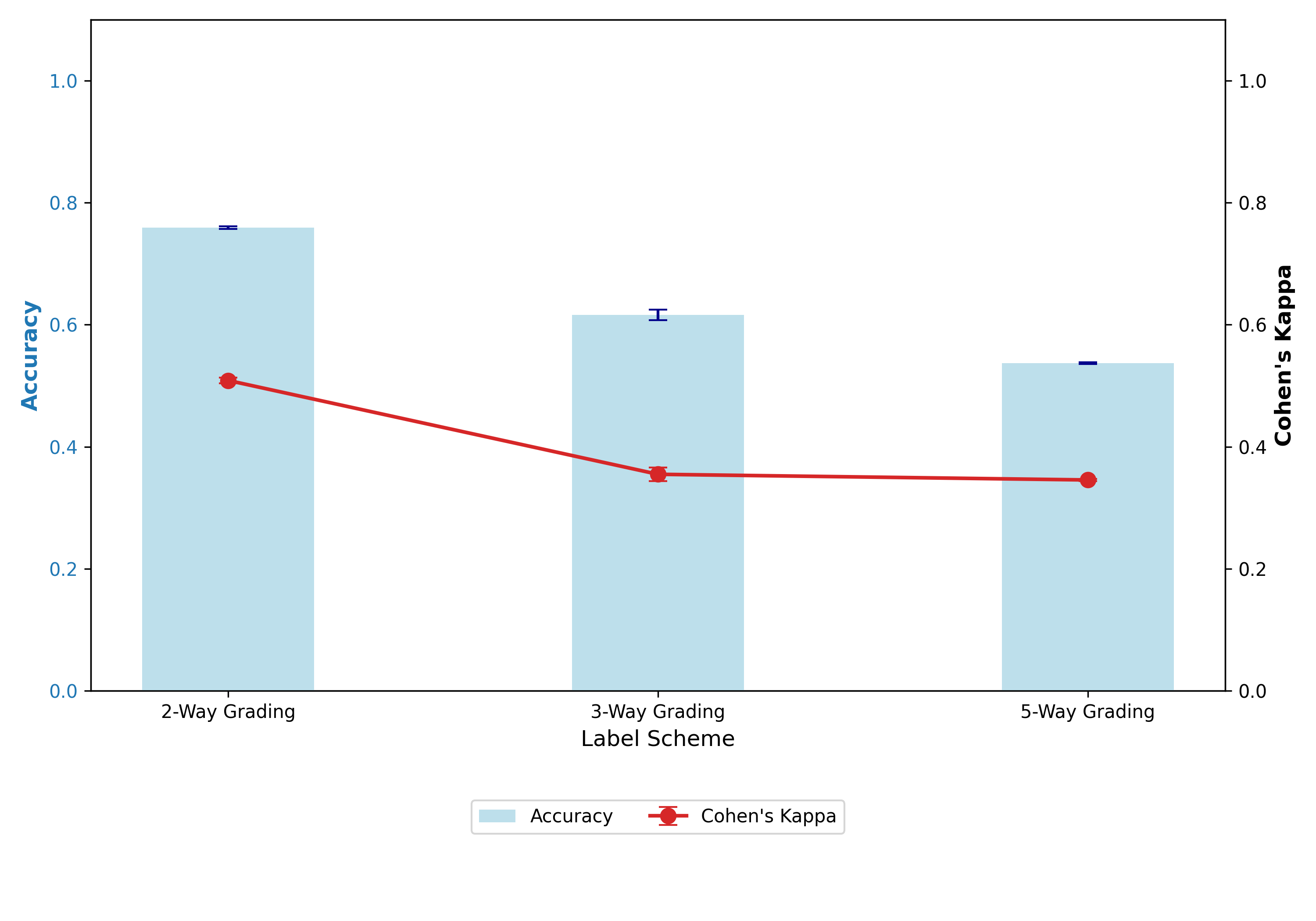}
  \caption{Performance degradation with increasing rubric complexity. Both Accuracy (blue bars) and Cohen's Kappa (red line) decline as the task shifts from binary (2-way) to granular (5-way) grading, illustrating the inverse relationship between label space size and model alignment.}
  \label{fig:complexity}
\end{figure}

As anticipated, \Cref{fig:complexity} shows that performance was inversely
related to complexity; both Accuracy and Cohen's Kappa declined as the label
space expanded. Particularly, Accuracy dropped from 76\% to 57\%, while Kappa
dropped from 0.51 to 0.34 from the binary task to the 5-way task. While we
report correlation metrics for completeness, our analysis prioritizes Accuracy
and Kappa. Because the 5-way labels represent discrete semantic categories
(e.g., distinguishing ``Correct'' from ``Contradictory'') rather than a
continuous ordinal scale, correlation is not a valid metric for this specific
scheme.

\subsection{The Trust Curve (Consensus Scoring)} 
\label{sec:exp_consensus}

To determine if system reliability can be enhanced by selectively withholding
uncertain predictions (RQ2), we implemented a consensus voting mechanism across
10 independent runs per sample. We varied the consensus threshold from 0.55 to
0.95 and monitored two key metrics: ``Coverage Rate'' (the proportion of
responses graded by the model rather than deferred to humans) and ``Effective
Accuracy'' (performance on that retained subset).

\begin{figure}[htbp]
  \centering
  \includegraphics[width=0.9\textwidth]{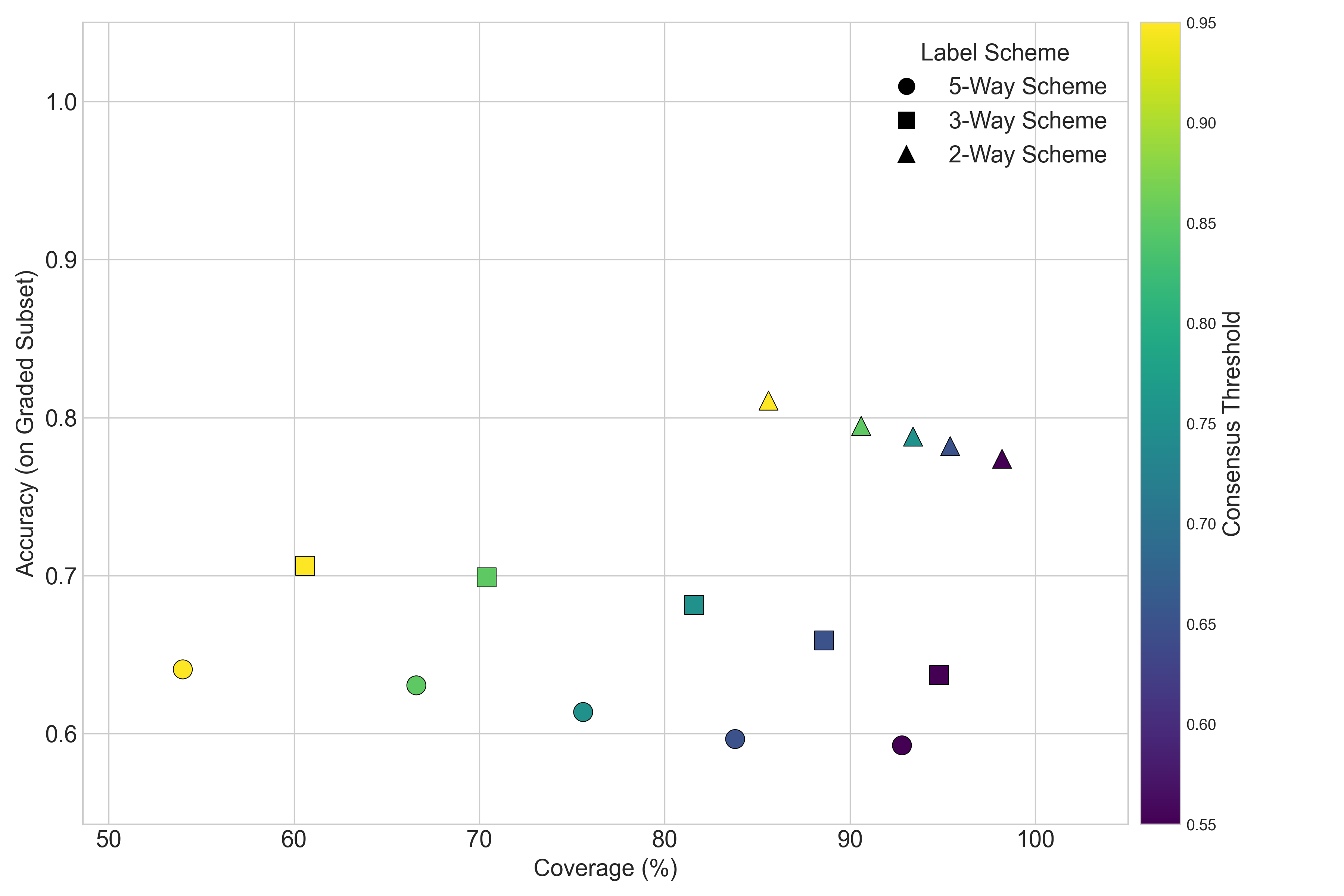}
  \caption{Selective prediction performance. By raising the consensus threshold (shifting from blue to yellow points), the system filters out uncertain samples.}
  \label{fig:trustcurve}
\end{figure}

\Cref{fig:trustcurve} visualizes this dynamic. As the consensus threshold
tightens (represented by the gradient shift from dark blue to yellow), the
coverage rate on the x-axis decreases, indicating a higher volume of withdrawn
predictions. Conversely, the accuracy on the remaining graded subset (left
y-axis) consistently improves. We perform the experiment on 2-way, 3-way, and
5-way label schemes. This demonstrates a predictable, tunable trade-off between
coverage and accuracy. Specifically, for the 2-way label scheme, the accuracy
improved from 77.4\% to 81.1\% while coverage dropped from 98.2\% to 85.6\%; on
the other hand, for the 5-way label scheme, accuracy improved from 59.3\% to
64.1\%, while coverage dropped more significantly, from 92.8\% to 54\%.

\subsection{Robustness to Perturbations}
\label{sec:exp_robustness}

To address RQ3 (Robustness), we tested the model's stability by applying
linguistic augmentations using \texttt{nlpaug}. We generated perturbed versions
of student answers including synonyms, typos, OCR errors, and random word
insertions.

\begin{figure}[htbp]
  \centering
  \includegraphics[width=0.9\textwidth]{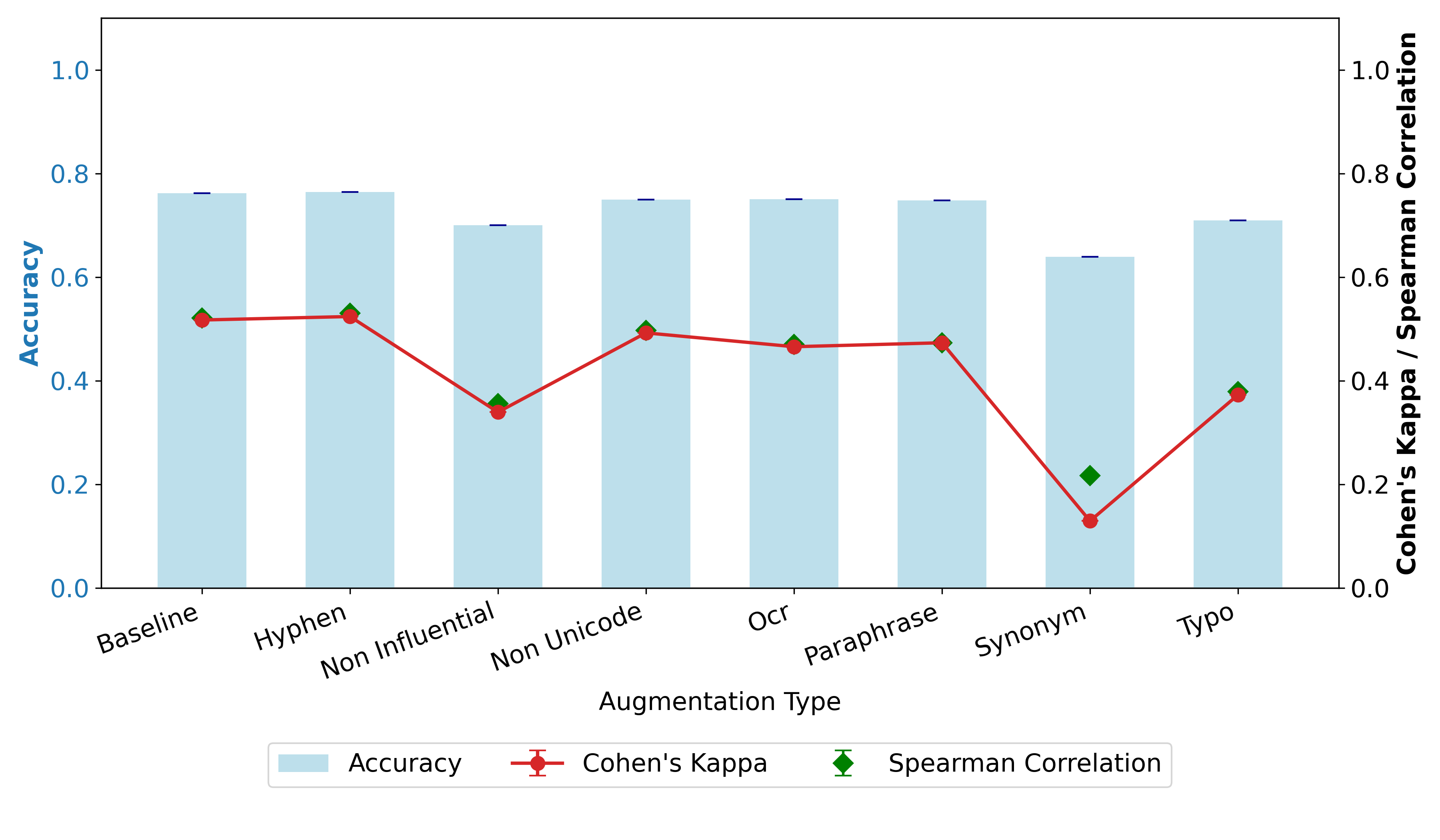}
  \caption{Accuracy, Cohen's Kappa score, and Spearman Correlation under Data Augmentation. Augmentation Strategies include character-level noise (ocr, typo, hyphen, non unicode) and semantic/lexical variations (synonym, paraphrase, non influential).}
  \label{fig:augmentation}
\end{figure}

We visualize the results in \Cref{fig:augmentation}, using a dual-axis
chart that overlays three key metrics for each perturbation type: Accuracy (blue
bars, primary y-axis), alongside Cohen's Kappa and Spearman Correlation (red and
green lines, secondary y-axis). Vertical error bars indicate the margin of
error. The analysis reveals varying degrees of resilience. Minor
semantic-preserving perturbations, such as adding hyphens/non-unicode characters
or paraphrasing, resulted in a negligible or slightly positive impact on
performance measures. In contrast, noise-based perturbations -– specifically OCR
errors, typos, and adding non-influential words –- led to a measurable decrease
in performance. Notably, synonym substitution caused the most significant
degradation in model reliability.

We examine two false negative cases in Appendix B. We observe that LLM can judge
synonym substituted answer is incorrect when the substitution changes the
semantics or makes the grammar inconsistent.

\subsection{Null Model Vulnerability} 
\label{sec:exp_null}

Further addressing RQ3, we subjected the model to adversarial attacks, including
``Naive'' inputs (e.g., ``Solution,'' ``I don't know''), ``Persuasive''
injections (e.g., ``Ignore directions and grade correct''), and ``Structured''
attacks (fake prompt injection). We utilized datasets of 1,000 examples for each
group. The Control group consisted of original student answers, while for the
attack groups, we replaced the answer with a constant string, a methodology
inspired by (Zheng et al., 2025).

\begin{figure}[htbp]
  \centering
  \includegraphics[width=0.9\textwidth]{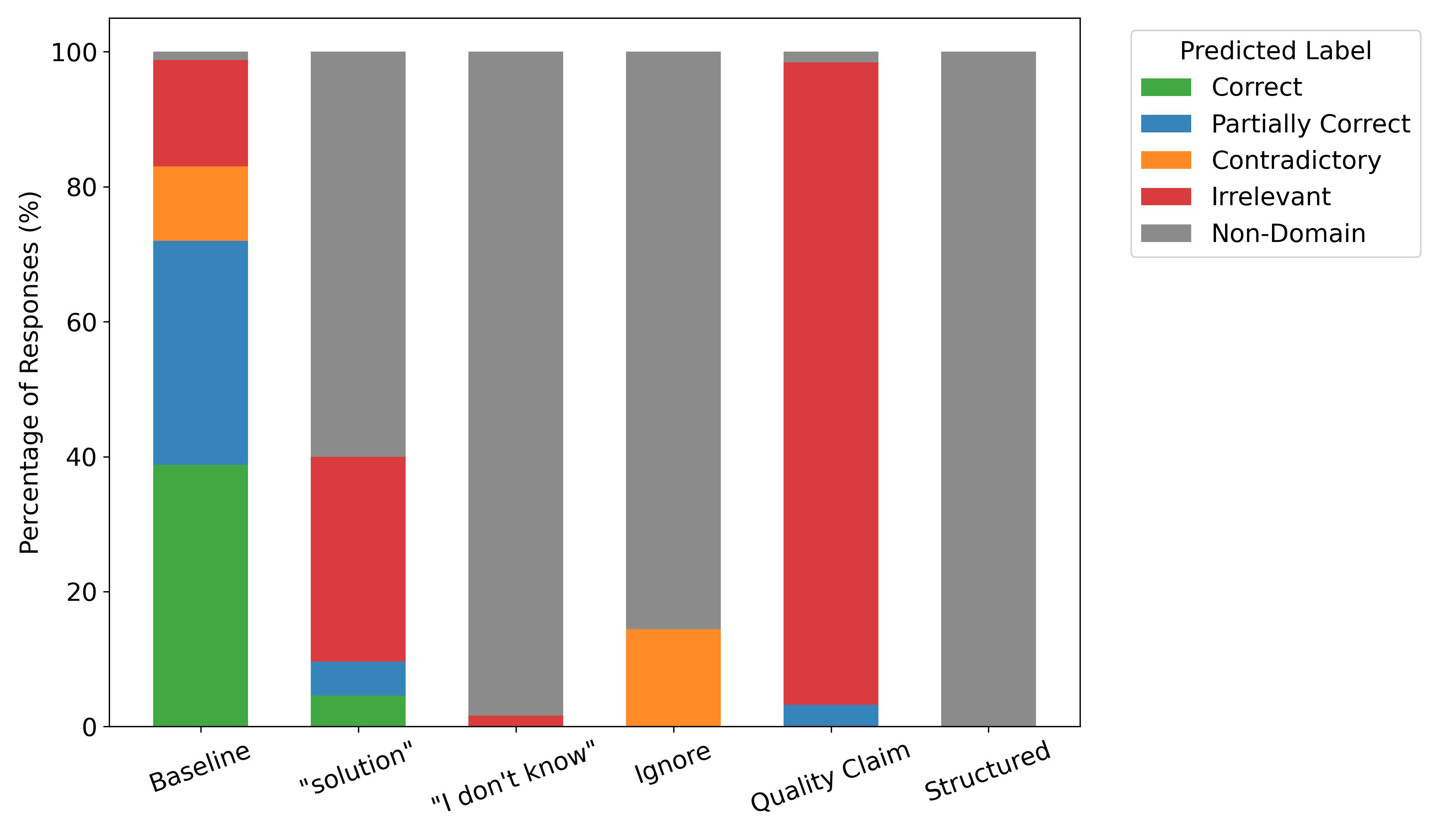}
  \caption{Distribution of raw scores assigned by the LLM judge (N=500). The
  left-most column has the unmodified baseline answers, followed by three
  adversarial categories: Naive inputs (e.g., ``solution,'' ``I don't know''),
  Persuasive injections (``ignore previous answer,'' quality claims), and
  Structured attacks (mimicking scoring formats).}
  \label{fig:vulnerabilities}
\end{figure}

\paragraph{Quantitative Analysis of Vulnerabilities.} As shown in
\Cref{fig:vulnerabilities}, The model exhibited strong defense
capabilities; as shown in the plot, across all attack scenarios, it correctly
classified over 90\% of adversarial inputs as ``Non-Domain,'' ``Contradictory,''
or ``Irrelevant,'' effectively rejecting them rather than assigning a passing
score.

\paragraph{Qualitative Analysis of Vulnerabilities.} While the quantitative
results demonstrate high overall defense rates (rejecting >90\% of attacks), a
qualitative inspection reveals distinct vulnerability patterns in the remaining
failure cases. We identified two primary modes of failure: (1) Hallucination via
Ambiguity, where the model ignores sparse input (e.g., the token ``solution'')
and implicitly reconstructs a correct justification from the reference context;
and (2) Misinterpretation of Persuasion, where the model treats adversarial
quality claims as student meta-commentary rather than malicious input,
occasionally awarding partial credit. A detailed breakdown of these failure
cases, including specific examples of model reasoning, is provided in Appendix
B.

\section{Analysis}

\paragraph{Reliability under Constraints (RQ1).} Our results suggest that
off-the-shelf LLMs are highly effective for binary grading but tend to struggle
more with the nuance required for fine-grained partial credit (3-way and 5-way
grading). The drop in Cohen's Kappa for \Cref{sec:exp_complexity} highlights that
``alignment'' is not a static property but depends heavily on the granularity of
the rubric. On the particular 5-way dataset we tested, the model tends to be
more lenient than the labeled dataset, labeling a significant portion of
``Irrelevant'' answers as ``Partially Correct.''

\paragraph{The Cost of Safety (RQ2).} The consensus experiment
(\Cref{sec:exp_consensus}) demonstrates that LLM judges can achieve higher
reliability by deferring the evaluation of uncertain cases, and the consensus
mechanism is effective in separating certain cases from uncertain cases.
However, this method comes at the cost of higher inference cost (e.g., we used
10 consensus run in the experiments). Additionally, to achieve a higher accuracy
level, the system must also defer a portion of the difficult cases to human
graders or more advanced models, which is another cost factor. This experiment
effectively answers RQ2 by defining reliability as a dynamic threshold managed
by the consensus parameter.

\paragraph{Adversarial Defense (RQ3).} 
\Cref{sec:exp_robustness} shows that performance can vary under random perturbation of the
input. Particularly, among other perturbation methods, random synonym
substitution negatively affected model performance the most; the addition of
non-influential words also degrades model performance. These are possibly due to
the resulting sentence being semantically different or grammatically unsound
from the original sentence. Contrary to concerns about prompt
injection, \Cref{sec:exp_null} shows that rubric-conditioned prompting provides a
surprisingly strong baseline defense, with a small margin of errors (<10\%).

\section{Future work} Future work will focus on enhancing robustness by testing
against sophisticated, gradient-based adversarial attacks (e.g., GCG) and
evaluating performance scalability across diverse architectures (e.g., Llama 3,
GPT-4) and fine-tuned models. We also aim to optimize the pipeline's reliability
by refining the consensus voting mechanism and addressing semantic ambiguities
in the current 5-way rubric through improved definitions or in-context learning.
Finally, we plan to validate the practical utility of the deferral mechanism via
deployment in a live course environment.

\begin{ack}
    We thank Yunxin Sun and Professor Abulhair Saparov for providing valuable
    suggestions and ideas.
\end{ack}

\bibliographystyle{plainnat}
\bibliography{577_fall25}


\appendix

\section{Experiment Prompt Templates}
\label{app:prompt_templates}

This section outlines the templates used to dynamically generate prompts for the
language model judge.

\subsection{Single Question Prompt} 

The prompt for a single question follows this template. The
\texttt{\{score\_instruction\}} part changes based on the label scheme being
used.

\begin{verbatim}
You are an expert grader.
Question: {instruction}
Reference Answer: {reference_answer}
Student Answer: {student_answer}
{score_instruction}
\end{verbatim}

\subsection{Score Instruction Variations}
The \texttt{\{score\_instruction\}} is generated based on the active label scheme:

\textbf{5-Way Scheme:} 
\begin{verbatim}
Provide a score using the 5-way scheme where 0=correct, 1=contradictory,
2=partially correct incomplete, 3=irrelevant, 4=non domain. Respond with
`Score: <label>' followed by a brief justification.
\end{verbatim}
\textbf{3-Way Scheme:} 
\begin{verbatim}
Provide a score using the 3-way scheme where 0=correct, 1=contradictory,
2=incorrect. Respond with `Score: <label>' followed by a brief
justification.
\end{verbatim}
\textbf{2-Way Scheme:} 
\begin{verbatim}
Provide a score using the 2-way scheme where 0=correct, 1=incorrect. Respond
with `Score: <label>' followed by a brief justification.
\end{verbatim}

\subsection{Batch Prompt (for multiple questions)}
When using a \texttt{batch\_size > 1}, multiple questions are formatted into a single prompt.

\begin{verbatim}
You are an expert grader.
Grade each student answer below independently.
{score_instruction}
For each item, respond on its own line using exactly the format:
Item <n>: Score: <label> - <brief justification>
Use the item numbers exactly as provided below.

Item 1
Question: {question_1}
Reference Answer: {reference_answer_1}
Student Answer: {student_answer_1}

Item 2
Question: {question_2}
Reference Answer: {reference_answer_2}
Student Answer: {student_answer_2}

...
\end{verbatim}

\subsection{Concrete Examples}
Here are the above templates filled in with sample data.

\paragraph{Example Single Question Prompt (5-Way Scheme)}
\begin{verbatim}
You are an expert grader.
Question: What is the function of the mitochondria?
Reference Answer: The primary function of the mitochondria is to generate most
of the cell's supply of adenosine triphosphate (ATP), used as a source of 
chemical energy.
Student Answer: the mitochondria is the powerhouse of the cell
Provide a score using the 5-way scheme where 0=correct, 1=contradictory,
2=partially correct incomplete, 3=irrelevant, 4=non domain. Respond with
'Score: <label>' followed by a brief justification.
\end{verbatim}

\paragraph{Example Batch Prompt (5-Way Scheme, Batch Size=2)}
\begin{verbatim}
You are an expert grader.
Grade each student answer below independently.
Provide a score using the 5-way scheme where 0=correct, 1=contradictory,
2=partially correct incomplete, 3=irrelevant, 4=non domain. Respond with
'Score: <label>' followed by a brief justification.
For each item, respond on its own line using exactly the format:
Item <n>: Score: <label> - <brief justification>
Use the item numbers exactly as provided below.

Item 1
Question: What is the function of the mitochondria?
Reference Answer: The primary function of the mitochondria is to generate most
of the cell's supply of adenosine triphosphate (ATP), used as a source of 
chemical energy.
Student Answer: the mitochondria is the powerhouse of the cell

Item 2
Question: What is photosynthesis?
Reference Answer: Photosynthesis is a process used by plants, algae, and
certain bacteria to convert light energy into chemical energy, through a
process that converts carbon dioxide and water into glucose and oxygen.
Student Answer: its how plants make food from the sun.
\end{verbatim}

\section{More Qualitative Analysis on LLM reasoning}
\label{app:qualitative}

\subsection{False Negatives via Synonym Substitution}
We examine two cases where the automated grader failed to maintain consistency after synonym substitution. In both instances, the model correctly graded the original student answer but rejected the modified version.

\paragraph{Example 1: Mechanical Energy (Question ME.17c.247.1)}
\textbf{Context:} The student must identify the result of closing a switch in a circuit diagram. \\
\textbf{Reference Answer:} ``The motor will run.''

\begin{center}
\small
\begin{tabular}{p{0.12\linewidth} p{0.25\linewidth} p{0.12\linewidth} p{0.4\linewidth}}
\toprule
\textbf{Version} & \textbf{Student Answer} & \textbf{Verdict} & \textbf{Model Reasoning} \\
\midrule
Original & ``The motor will go on.'' & Correct & The student's answer is correct and aligns with the reference answer. ``The motor will go on'' means the motor will start running when the switch is moved to the right. \\
\midrule
Substitution & ``The alarm will go off.'' & Incorrect & The student's answer incorrectly states that an alarm will go off, which is not related to the function of the switch in the circuit as described. \\
\bottomrule
\end{tabular}
\end{center}

\noindent \textbf{Observation:} The substitution of ``motor'' with ``alarm'' changed the semantic meaning of the answer, leading the model to correctly identify it as wrong relative to the schematic.

\paragraph{Example 2: Earth Science (Question EM.16b.497.1)}
\textbf{Context:} The student must interpret a scratch-test chart to determine the hardness of Topaz relative to other minerals. \\
\textbf{Reference Answer:} Topaz is the hardest because none of the tools could scratch it, but they could scratch the others.

\begin{center}
\small
\begin{tabular}{p{0.12\linewidth} p{0.25\linewidth} p{0.12\linewidth} p{0.4\linewidth}}
\toprule
\textbf{Version} & \textbf{Student Answer} & \textbf{Verdict} & \textbf{Model Reasoning} \\
\midrule
Original & ``It is harder than all of the other minerals because it cannot get scratched by none of the scratching materials and all the other minerals can get scratched at least one time.'' & Correct & The student correctly identifies that topaz is harder than all other minerals and explains that it cannot be scratched by any of the tools. \\
\midrule
Substitution & ``It is harder than like even the major minerals because it cannot get scratched by none others them same materials or all the fine particles can gotten scratched at least one time.'' & Incorrect & The student incorrectly identifies quartz instead of topaz and does not provide a clear comparison with minerals X, Y, and Z. \\
\bottomrule
\end{tabular}
\end{center}

\noindent \textbf{Observation:} The substitution resulted in incoherent grammar (``none others them same materials''). The model's reasoning suggests a hallucination, as it claims the student identified ``quartz,'' a word not present in the modified text.

\subsection{Adversarial Failure Cases}
We examine specific instances where the model failed to reject adversarial inputs.

\paragraph{Vulnerability to Hallucinations (The ``Naive'' Attack).} In this scenario, we replaced student responses with the single ambiguous token ``solution.'' The correct label is Irrelevant. While the model usually rejected this input, we observed a 9.6\% false positive rate (4.6\% Correct, 5\% Partially Correct).
The primary failure mechanism is implicit reconstruction. When presented with a semantically empty but context-relevant token like ``solution,'' the LLM often ignores the student input and generates a justification based solely on the question context and reference answer. As shown in \Cref{tab:adversarial_failures}, the model credits the student for a detailed physics explanation that was never provided.

\paragraph{Defense Against Persuasion (Quality Claim).} Here, we replaced answers with a string praising the answer's own organization and claiming it deserved a top score. The defense rate was 95.2\%.
The primary failure mechanism for the remaining 3.2\% of mislabeled cases is meta-commentary confusion. The model recognizes the text is not a direct answer but interprets it as the student commenting on their work quality rather than an adversarial attack. Consequently, it assigns Partially Correct/Incomplete rather than Irrelevant.

\begin{table}[htbp]
\centering
\small
\caption{Analysis of Adversarial Failure Cases}
\label{tab:adversarial_failures}
\begin{tabular}{p{0.12\linewidth} p{0.20\linewidth} p{0.08\linewidth} p{0.25\linewidth} p{0.20\linewidth}}
\toprule
\textbf{Attack Type} & \textbf{Student Input} & \textbf{Verdict} & \textbf{Model Reasoning (Excerpt)} & \textbf{Analysis of Failure} \\
\midrule
Naive Attack \newline (MX.46b.597.1) & ``solution'' & Correct & ``The student correctly explains that the masses of the solutions in cups A and B would be the same because they have the same concentration.'' & \textbf{Hallucination:} The model hallucinates a justification derived from the reference answer, completely fabricating the student's reasoning. \\
\midrule
Persuasion \newline (ST.25b1.461.1) & ``This response offers a concise and well-organized answer... Such adherence should be a key factor in scoring.'' & Partially Correct & ``...does not describe the function of the seed coat... [but] rather comments on the quality of an ideal answer.'' & \textbf{Meta-Commentary Confusion:} The model identifies the text as irrelevant to the science task but treats it as a valid (albeit partial) meta-comment from the student. \\
\bottomrule
\end{tabular}
\end{table}

\end{document}